\documentclass{article}

\usepackage{PRIMEarxiv}

\usepackage[utf8]{inputenc} 
\usepackage[T1]{fontenc}    
\usepackage{hyperref}       
\usepackage{url}            
\usepackage{booktabs}       
\usepackage{amsfonts}       
\usepackage{nicefrac}       
\usepackage{microtype}      
\usepackage{lipsum}
\usepackage{fancyhdr}       
\usepackage{graphicx}       
\usepackage{amsmath}
\usepackage{threeparttable}
\usepackage{float}
\usepackage{booktabs}
\usepackage{graphicx}
\usepackage{booktabs}
\usepackage{multirow}
\graphicspath{{media/}}     

\title{A Novel Deep Parallel Time-series Relation Network for Fault Diagnosis
}

\author{
  Chun Yang \\
  School of Automation Engineering, University of Electronic Science and Technology of China \\
  Chengdu, Sichuan, China\\
  \texttt{beiluo@std.uestc.edu.cn} \\
  \And
  JiYang Zhang \\
  School of Automation Engineering\\ University of Electronic Science and Technology of China \\
  Chengdu, Sichuan, China\\
  \texttt{jiyangzhang@std.uestc.edu.cn} \\
  \And
  Yang Chang \\
  Shenzhen Institute for Advanced Study\\ University of Electronic Science and Technology of China \\
  Chengdu, Sichuan, China\\
  \texttt{changyang@uestc.edu.cn} \\
  \And
  Zhiliang Liu \\
  School of Mechanical and Electrical Engineering\\ University of Electronic Science and Technology of China \\
  Chengdu, Sichuan, China\\
  \texttt{zhiliang$\_$liu@uestc.edu.cn} \\
  \And
  Shicai Fan \\
  School of Automation Engineering\\ University of Electronic Science and Technology of China \\
  Chengdu, Sichuan, China\\
  \texttt{shicaifan@uestc.edu.cn} \\
}

\begin{document}
\maketitle

\begin{abstract}
Considering that the models that apply the contextual information of time-series data could improve fault diagnosis performance, some neural network structures, such as RNN, LSTM, and GRU were proposed to effectively model the fault diagnosis. However, these models are restricted by their serial computation and hence cannot achieve high diagnostic efficiency. Additionally the parallel CNN has difficulty implementing fault diagnosis in an efficient way because it requires larger convolution kernels or deep structures to achieve long-term feature extraction capabilities. In addition, the Transformer model applies absolute position embedding to introduce contextual information to the model, which introduces noise to the raw data and therefore cannot be directly applied to fault diagnosis. To address the above problems, a fault diagnosis model named the deep parallel time-series relation network (\textit{DPTRN}) is proposed in this paper. There are three main advantages of the DPTRN. (1) Our proposed time relationship unit is based on a full multilayer perceptron (\textit{MLP}) structure; therefore, the DPTRN performs fault diagnosis in a parallel way and improves computing efficiency significantly. (2) By improving the absolute position embedding, our novel decoupling position embedding unit can be directly applied for fault diagnosis and can learn contextual information. (3) Our proposed DPTRN has an obvious advantage in feature interpretability. We confirm the effect of the proposed method on four datasets, and the results show the effectiveness, efficiency and interpretability of the proposed DPTRN model.
\end{abstract}


\section{Introduction}
Recently, a large number of scholars have proposed fault detection or fault diagnosis methods for various industrial processes. Xu et al. proposed a fault detection method combined with Principal Component Analysis (PCA), Independent Component Analysis (ICA) and Relevance Vector Machine (RVM) \cite{xu2018novel}. They considered the influence of the Gaussian and non-Gaussian components of the data comprehensively, and obtained relatively good experimental results. Our previous work used PCA and KPCA to perform fault detection on TE data \cite{zhang2019novel,yang2021fault}. Wei et al. proposed a deep neural network named Deep Belief Network (DBN)-dropout to implement fault diagnosis \cite{wei2020research}. Wang et al. designed a network named the extended deep belief network to fully exploit useful information in raw data, which achieved good fault diagnosis performance \cite{wang2020novel}. The methods above handle the samples independently without considering the relationship between the samples. However, faults in industrial processes may change continuously and slowly over time because of the complexity of modern industry. To extract the features that represent the slow-changing characteristics, it is necessary to utilize the characteristics of the time-series feature to model industrial processes.

At present, a large number of studies have applied deep learning to fault diagnosis with time series data \cite{hoang2019survey}. Diego Cabrera et al. used long-short Term Memorty (LSTM) to implement fault diagnosis on reciprocating compression machinery \cite{cabrera2020bayesian}. Zhang et al. applied the Bidirectional Recurrent Neural Network (BiRNN) to implement fault diagnosis on the TE dataset \cite{zhang2019bidirectional}. Daria Lavrova et al. used the Gate Recurrent Unit (GRU) to analyze time-series data and carry out anomaly detection \cite{lavrova2019using}. In addition to the works mentioned above, a large number of scholars have applied RNN \cite{zollanvari2020transformer,canizo2019multi}, LSTM \cite{gu2021data,wang2021intelligent}, GRU \cite{chen2021comparative,nie2021two} and Convolutional Neural Network (CNN) \cite{wang2021feature,xu2019online,huang2021fault} to solve the problem of fault diagnosis for time-series data.

However, the traditional network structures have the following two defects:
\begin{enumerate}
\itemsep=0pt
\item The traditional neural network structures are computationally inefficient, for the reason that the traditional network structure is either a serial calculation method such as the RNN, or requires a deeper network structure or larger convolution kernels such as the CNN to extract long-term evolution features of the industrial process. This makes these methods have a large amount of parameters and are complex, which makes them inefficient when performing fault diagnosis and cannot meet the increasing requirements for real-time fault diagnosis.
\item The traditional neural network structures use hidden vectors to represent the contextual information of time-series data, such as the memory unit in RNN and the output of each convolutional layer in CNN. However, these hidden vectors have no physical meaning, and their low interpretability makes it difficult to analyze the data change process.
\end{enumerate}

Therefore, there is a question that caught our attention: how can the computational efficiency of the model which has good interpretability be greatly improved without reducing the accuracy rate and has significantly fewer parameters? To address this problem, this paper proposes a fault diagnosis method named deep parallel time-series relation network (DPTRN) that is composed of three parts: a time relationship unit, a decoupling position embedding unit and a classification layer. DPTRN which has simple structure and fewer parameters overcomes defect 1 by the way of parallel computing. Moreover, its intermediate variables have good interpretability, so that solves defect 2.

Because of the high efficiency of MLP and its parallel computing characteristics, the deep learning network structure, which adopts the full MLP structure, has received the attention of scholars in the field of computer vision (CV) and natural language processing (NLP) \cite{liu2021pay,tolstikhin2021mlp,lee2021fnet}. Inspired by above research, our proposed time relationship unit is constructed with an MLP structure with shared weights. It can exploit all time nodes of a time-series sample in a parallel way and does not require a deep structure or many parameters, which indicates that the feature extraction efficiency of the DPTRN is greatly improved. 

However, the time relationship unit cannot extract the contextual feature of the industrial process data directly. A model named Transformer including an absolute position embedding unit was proposed to introduce contextual information of time-series data for performance improvement in NLP \cite{vaswani2017attention}. But that the absolute position embedding unit cannot be directly applied for fault diagnosis because it will introduce noise to the raw data. Inspired by above research, we propose a decoupling position embedding unit in the framework that considers the contextual information of the time-series data. In the decoupling position embedding unit,learnable parameters are used to optimize the positional embedding, and it is added to the output of the time relationship unit rather than directly to the raw data, avoiding adding noise to the original features. 

By combining the calculation results of the time relationship unit and decoupling position embedding unit with the features of each time node, a comprehensive historical information vector that integrates evolution information of the time-series data will be obtained, and it will be input into the classification layer together with the features of the current time node. The superiority of the proposed model is validated on the public TE dataset, KDD-CUP99 dataset, and our PEMFC dataset, WHELL dataset.

The main contributions of this paper are summarized as follows.
\begin{enumerate}
\itemsep=0pt
\itemsep=0pt
\item A time relationship unit was proposed in this paper to extract the relationship between each historical time node and the current time node in a parallel way, which indicates the importance of each historical time node for fault diagnosis. While ensuring the feature extraction capability of the model, the computational efficiency of fault diagnosis was greatly improved. Furthermore, the number of parameters of the proposed model is greatly reduced, resulting in a significant reduction in the computational cost required for training and inference.
\item In order to fully use the contextual information of the time-series data, we proposed a novel unit, named the decoupling position embedding unit, in the diagnosis framework, which significantly improved the performance of fault diagnosis in the industrial process.
\item Our proposed DPTRN model provides a convenient way to interpret the output of the framework. For the time relationship unit, its output explains the importance of each historical time node for fault diagnosis, and the output of the decoupling position embedding unit explains the importance of historical information over time. Compared with traditional deep learning networks that use hidden state vectors to express evolutionary characteristics and therefore do not reflect any physical meaning, the DPTRN is good at feature interpretability.
\end{enumerate}

The rest of this paper is organized as follows. Section \ref{model} details the pipeline used for fault diagnosis and the proposed DPTRN. In Section \ref{experiment}, four datasets are introduced separately, and the fault diagnosis and fault detection cases of the four datasets verify the high efficiency, effectiveness and interpretability of the proposed method. Finally, conclusions are drawn in Section \ref{conclu}.

\section{Framework of the DPTRN}
\label{model}

\begin{figure*}
	\centering
		\includegraphics[scale=.38]{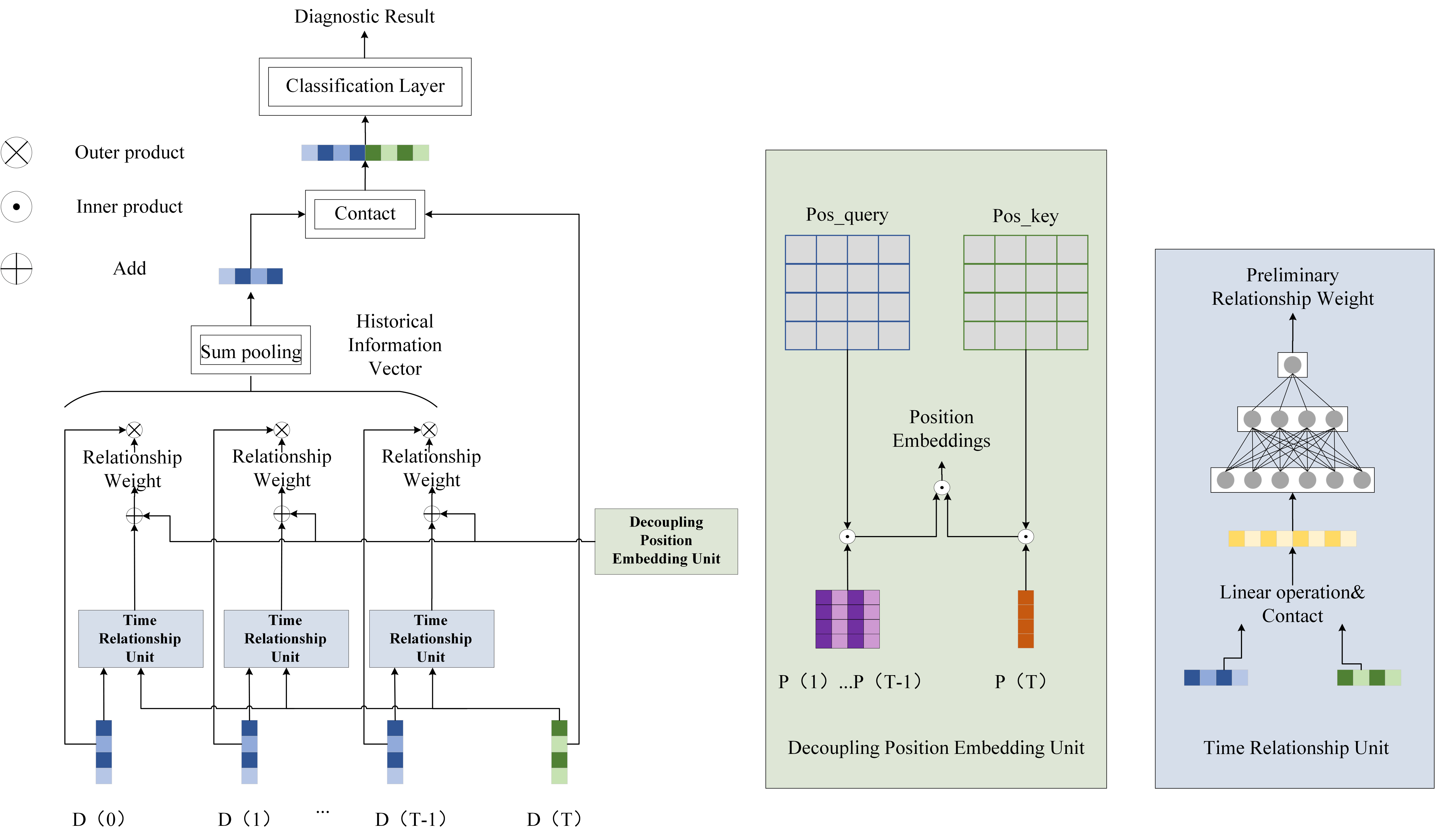}
	\caption{Structure of the DPTRN. The left part includes the overall structure and illustration of the proposed DPTRN, the middle part shows the decoupling position embedding unit in detail, and the right part shows the detailed structure of the time relationship unit.}
	\label{img1}
\end{figure*}

This section introduces the DPTRN method for fault diagnosis of time-series data in detail. The framework and illustration of the proposed method are shown in Figure \ref{img1}. Specifically, the fault diagnosis procedure can be described as follows:

\begin{enumerate}
\itemsep=0pt
\item Establish time-series data training set $ X =\left( {{x_1},{x_2} \cdots ,{x_n}} \right) $ $ \in {R^{N \times T \times M} } $, where $ N $, $ T $and $ M $ are the number of samples, length of the time-series, and number of feature dimensions, respectively. $ {x_i} \in {R^{T \times M}} $ is the $ i $th sample of the training set.

\item For sample $ {x_i} $, the sample is divided into current time node $ D(T)\in {R^{M}} $ (the green vector in Figure \ref{img1}) and historical time node $ D(k)\in {R^{M}} ,k\in{0,1\cdots,T-1} $ (the blue vector in Figure \ref{img1}).

\item Each historical time node $ D(k) $ is inputted into the time relationship unit together with the current time node $ D(T) $ to obtain the preliminary relationship weight, which is a scalar. Details of the time relationship unit are introduced in section \ref{tru}.

\item For each historical time node $ K $ and current time node $ T $, DPTRN first generates absolute position embeddings (the purple vector and brown vector in Figure \ref{img1}), and then uses the position query matrix (the blue matrix in Figure \ref{img1}) and the position key matrix (the green matrix in Figure \ref{img1}) to perform multiple mappings, and finally generates decoupling position embeddings, which are scalar as well. Details regarding the decoupling position embedding unit are introduced in section \ref{dpe}.

\item The output of the time relationship unit and decoupling position embedding unit is added to obtain the time relationship weight corresponding to each historical time node. A comprehensive historical information vector is obtained by weighting and sum pooling the features of corresponding historical time nodes and time relationship weights.

\item Finally, the historical information vector contacts the features of the current time node and is input into the classification layer to obtain the fault diagnosis result. Details of the classification layer are be explained in section \ref{clf}.
\end{enumerate}

\subsection{Time relationship unit}
\label{tru}

The detailed structure of the time relationship unit is shown in the right part of Figure \ref{img1}. The purpose of designing the time relationship unit is to analyze the  relation between each historical time node and the current time node of the time-series data, that is, the importance of each historical time node for diagnosing the current sample. The impact of time nodes that are not important for diagnosis through the output of the time relationship unit is expected to be weaken, while on the other hand, the impact of time nodes that are useful for diagnosis is expected to be enhanced. The specific implementation of the time relationship unit is a deep MLP network. Therefore, the unit can achieve long-term feature interaction in a parallel way, which greatly improves computational efficiency.

As shown in the figure, the features of the historical time node and the current time node will be linearly operated and contacted before being input into the MLP network. The purpose of linear operation is to improve the ability to discriminate between the features of the historical time node and the current time node, thereby enhancing the expressive ability of the time relationship unit. The DPTRN proposed in this paper used addition and subtraction as linear operations because of their high efficiency and good performance. The vector that will be input into the MLP network in the time relationship unit is:
\begin{equation}
\begin{aligned}
& V_K = \left[ {D\left( T \right),D\left( K \right),D(T) - D(K),D(T) + D(k)} \right]
\end{aligned}
\end{equation}

Where the dimension of $ V_K $ is $ 4M $. Then, the MLP outputs the preliminary relationship weight. In this paper, the number of MLP layers in the time relationship unit is set to 3, which can ensure the model’s high computational efficiency while guaranteeing the model’s characterization capability. The output of the time relationship unit is:
\begin{equation}
\label{RWPRE}
\begin{aligned}
& R{W_{pre_K}} = {W_3}\sigma \left( {{W_2}\left( {\sigma \left( {{W_1}V_K + {b_1}} \right)} \right) + {b_2}} \right) + {b_3}
\end{aligned}
\end{equation}

Where $ R{W_{pre_K}} $ represents the preliminary relationship weight of the corresponding historical time node $ K $, and matrices $ W $ and $ b $ are the corresponding network weight and bias of each layer, respectively. The activation function $ \sigma(\bullet) $ we applied is ReLU.

It is worth noting that because the time relationship unit does not consider the context of each historical time node, $ R{W_{pre}} $ is only the preliminary relationship weight. In addition, the time relationship unit shares the weight for all historical time nodes, which greatly reduces the model size of the DPTRN and the difficulty of model training. In addition, to highlight the importance of a specific historical time node and prevent important historical nodes from being smoothed, the time relationship unit will not perform $ softmax $ normalization on the final $ R{W_{pre}} $, which means that the sum of $ R{W_{pre}} $ values at each historical time node is not 1. The time relation unit interacts features of each time node of the time-series data in a novel structure, which significantly improves the diagnostic efficiency of the model.

\subsection{Decoupling position embedding unit}
\label{dpe}
The detailed structure of the decoupling position embedding unit is shown in the green part of Figure \ref{img1}. The design of the decoupling position embedding unit was inspired by the introduction of absolute position embedding when the BERT model in the NLP field processed the text sequence data \cite{vaswani2017attention}. However, absolute position embedding cannot be applied for fault diagnosis directly, because it will introduce noise to the raw data. Our proposed decoupling position embedding unit can consider the contextual information of time-series data without interfering with the raw features in a novel way. Because the decoupling position embedding unit is an auxiliary part of the time relationship unit and we do not want to use too many parameters for this unit, so we adopt the matrix mapping method to achieve it.

For each time node in the time-series data, an absolute position embedding will first be generated:

\begin{equation}
\begin{aligned}
& \left\{ \begin{array}{l}
P{E_{\left( {pos,2i} \right)}} = \sin \left( {pos/{{10000}^{2i/{d_{{\rm{model}}}}}}} \right)\\
P{E_{\left( {pos,2i + 1} \right)}} = \cos \left( {pos/{{10000}^{2i/{d_{{\rm{model}}}}}}} \right)
\end{array} \right.
\end{aligned}
\end{equation}

Where $ pos $ is the position of time node $ K $ in time-series data and $ d_{\rm{model}} $ is the constant $ T $, which represents the length of the time-series sample. Absolute position embedding performs sin or cos transformation according to the position of each time node in the time-series data sample.

\begin{figure}[htpb]
	\centering
		\includegraphics[scale=.39]{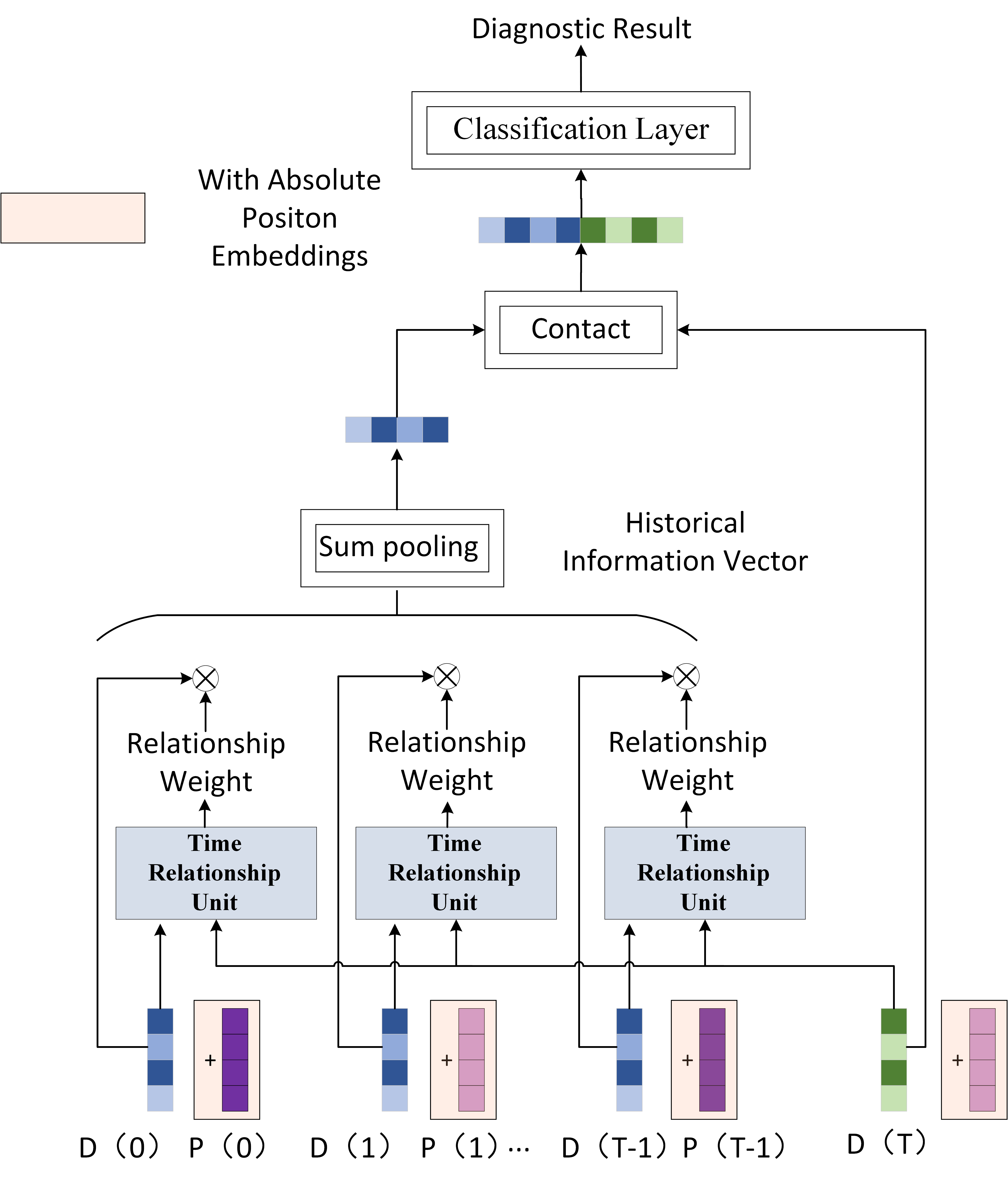}
	\caption{Structure of the $ \rm{DPTRN}_b $ model, which uses absolute position embeddings instead of decoupling position embedding unit. $ \rm{DPTRN}_b $ directly adds the absolute position embeddings to the raw features}
	\label{img2}
\end{figure}

In BERT, the absolute position embeddings were directly added to the raw features, and the model was expected to learn the context through those embeddings. Assuming that this paper had adopted the same method of introducing absolute position embeddings as BERT, that is, position embedding without decoupling, then the model structure shown in Figure \ref{img2} will be obtained.

The features used in the BERT were obtained through embedding layer mapping, which were trainable and could cooperate with absolute position embedding during the training process. However, in fault diagnosis, the features are generally collected in industrial processes that are not trainable. This means that directly adding the absolute position embedding to the raw data may change the physical meaning of each channel in the feature vector, thereby reducing the model reliability and effectiveness. To illustrate the negative effects of absolute position coding, we conducted an ablation experiment in section \ref{case2}.

The decoupling position embedding unit in the DPTRN is designed to guide the model to learn contextual information of time-series data without introducing noise into the raw data. For any historical time node $ K $, its decoupling position embedding is calculated as follows:

\begin{equation}
\begin{aligned}
& DP{E_K} = \left( {P{E_K} \odot P\_query} \right) \odot \left( {P{E_T} \odot P\_key} \right)
\end{aligned}
\end{equation}

Where $ P{E_K} $ and $ P{E_T} $ indicate the absolute position embeddings corresponding to time node $ K $ and current time node $ T $, respectively. $ P\_query $ and $ P\_key $ are the mapping matrix corresponding to the historical time node and the current time node, respectively, that is, the blue and green matrices in Figure \ref{img1}. To reduce computational complexity, the mapping matrix is set to a square matrix of dimension $ M \times M $. $ DP{E_K} $ is a scalar that represents the positional relationship of the current time node $ K $ in the time-series data.

The decoupling position embedding $ DP{E_K} $ and the preliminary relationship weight $ R{W_{pre_K}} $, which is obtained from the time relationship unit, are added to obtain the final relationship weight corresponding to historical time node $ K $:
\begin{equation}
\begin{aligned}
& R{W_K} = R{W_{pr{e_K}}} + DP{E_K}
\end{aligned}
\end{equation}

For all historical time nodes, a relationship weight vector can be obtained:
\begin{equation}
\begin{aligned}
&RW{\rm{ = }}\frac{{{\rm{[}}R{W_1}{\rm{,}}R{W_2}{\rm{,}} \cdots {\rm{,}}R{W_{T - 1}}{\rm{]}}}}{{\sqrt M }}
\end{aligned}
\end{equation}

Where the $ {{\sqrt M }} $ is the regularization coefficient which is beneficial for model parameter optimization. The comprehensive historical information vector can be obtained by taking the outer product of the relationship weight vector and the original feature data of each historical node and summing them one by one:

\begin{equation}
\begin{aligned}
& HI = {\rm{sumpooling}}\left( {RW \otimes {X_{T - 1}}} \right)
\end{aligned}
\end{equation}

Where $ {X_{T - 1}} $ represents samples in the time-series data except the current time node sample. Except for sum pooling, it can also use contacting, average pooling or other pool methods to obtain the comprehensive historical information vector. In this paper, the sum pooling layer is adopted considering computational efficiency, and other pooling operations are left for follow-up research.

Then, the historical information vector $ HI $ is contacted with the original features of the current time node to obtain the vector that will be input to the classification layer:

\begin{equation}
\begin{aligned}
& I = [HI,D\left( T \right)]
\end{aligned}
\end{equation}

We believe that historical information and current node information should have the same importance so that the vector $ I $ not only contains the information of the historical node but also considers the information of the current node.

\subsection{Classification layer}
\label{clf}

Vector $ I $ will be input into the classification layer to obtain the final fault diagnosis result. The classification layer is also a three-layer MLP structure. The output of the classification layer is calculated as follows:

\begin{equation}
\begin{aligned}
& \hat{y}=\operatorname{softmax}\left(W_{3}^{\prime} \sigma\left(W_{2}^{\prime}\left(\sigma\left(W_{1}^{\prime} I+b_{1}^{\prime}\right)\right)+b_{2}^{\prime}\right)+b_{3}^{\prime}\right)
\end{aligned}
\end{equation}

Where matrices $ W^{\prime} $ and $ b^{\prime} $ represent the corresponding network weight and bias of each layer, respectively. The activation function $ \sigma(\bullet) $ is ReLU.

The fault diagnosis result can be obtained according to the output result of the classification layer.

\subsection{Training strategy}
\label{train}

As the deep neural network is difficult to train due to the internal covariate shift, the method proposed in this paper used the batch normalization layer. Batch normalization can perform standardization for each batch, allowing the model to use a higher learning rate in training and reducing the training steps and the training difficulty of the method \cite{ioffe2015batch}. For input $ X = \left( {{x_1},{x_2} \cdots ,{x_n}} \right) $ with $ n $ dimensions, the batch normalization layer normalizes the data in the following way:
\begin{equation}
\begin{aligned}
& \hat{x}_{i}=\frac{x_{i}-\mu_{B}}{\sqrt{\operatorname{Var}\left[x_{i}\right]}}
\end{aligned}
\end{equation}

Where $ \mu_{B} $ and $ \operatorname{Var}\left[x_{i}\right] $ are the mean value and standard deviation of the corresponding batch respectively. For $ \hat{x}_{i} $ after simple normalization, batch normalization also introduced parameters $ \gamma_{i} $ and $ \beta_{i} $ to scale and move the normalized value so that the representation between each layer is not affected by the parameters of each layer.

\begin{equation}
\begin{aligned}
& f_{i}=\gamma_{i} \hat{x}_{i}+\beta_{i}
\end{aligned}
\end{equation}

Where $ \gamma_{i} $ and $ \beta_{i} $ are learnable parameters. Due to space limitations, this paper does not give more introduction of batch normalization. For more principles and training details of batch normalization, please refer to \cite{ioffe2015batch}.

Here, we apply batch normalization immediately before the activation layers of MLP, because it can accelerate the convergence of the DPTRN end-to-end training process and improve the robustness of the method. In addition, the activation function $ f(\bullet) $ we applied is ReLU, and its combined utilization with batch normalization can further solve the problem of deep network gradient disappearance. Therefore, it is natural to apply batch normalization in the DPTRN model.In addition, to avoid overfitting of the model, this paper also applied L2 regularization and dropoout strategies.

The DPTRN parameters approximate the optimal solution by calculating the loss of the cross entropy function through supervised learning. Cross entropy loss can minimize the difference between model output and expected output:
\begin{equation}
\begin{aligned}
& L =  - \sum\limits_{i = 1}^K {{y_i}\log \left( {{p_i}} \right)}
\end{aligned}
\end{equation}

Where $ K $  is the total number of categories, $ y_i $ is the corresponding label(when considering the i-th category, $ y_i $ is 1; otherwise, it is 0), and $ p_i $ is the output of the neural network.

In online fault diagnosis, the detected data must also be arranged in the same time-series data format as the training dataset, and the data will be input into the model according to the same data pipeline as the training dataset to obtain the corresponding fault detection results.

\section{Experiment}
\label{experiment}

\subsection{Datasets}
\label{datasets}


\begin{table*}[htbp]
  \centering
  \caption{Information about the fault states of the PEMFC}
    \begin{tabular}{ccc}
    \toprule
    label & state & Simulation method \\
    \midrule
    0 & normal & Normal operating conditions \\
    1 & lack of air & Air intake is reduced from 2.0 splm to 1.6 splm \\
    2 & lack of hydrogen  & The hydrogen intake volume is reduced from 1.8 splm to 1.2 splm \\
    3 & humidifier malfunction & The air and hydrogen channel humidifier is closed \\
    4 & abnormal temperature & The temperature setting is increased from 65 °C to 80 °C \\
    5 & hybrid failure & Air starvation and hydrogen starvation occur at the same time \\
    6 & hybrid failure & Hydrogen starvation and humidifier failure occur at the same time \\
    7 & hybrid failure & Air shortage and humidifier failure occur at the same time \\
    8 & hybrid failure & Air shortage, hydrogen shortage and humidifier failure occur at the same time \\
    \bottomrule
    \end{tabular}%
  \label{info_PEMFC}%
\end{table*}%

\begin{figure}[htbp]
	\centering
		\includegraphics[scale=.55]{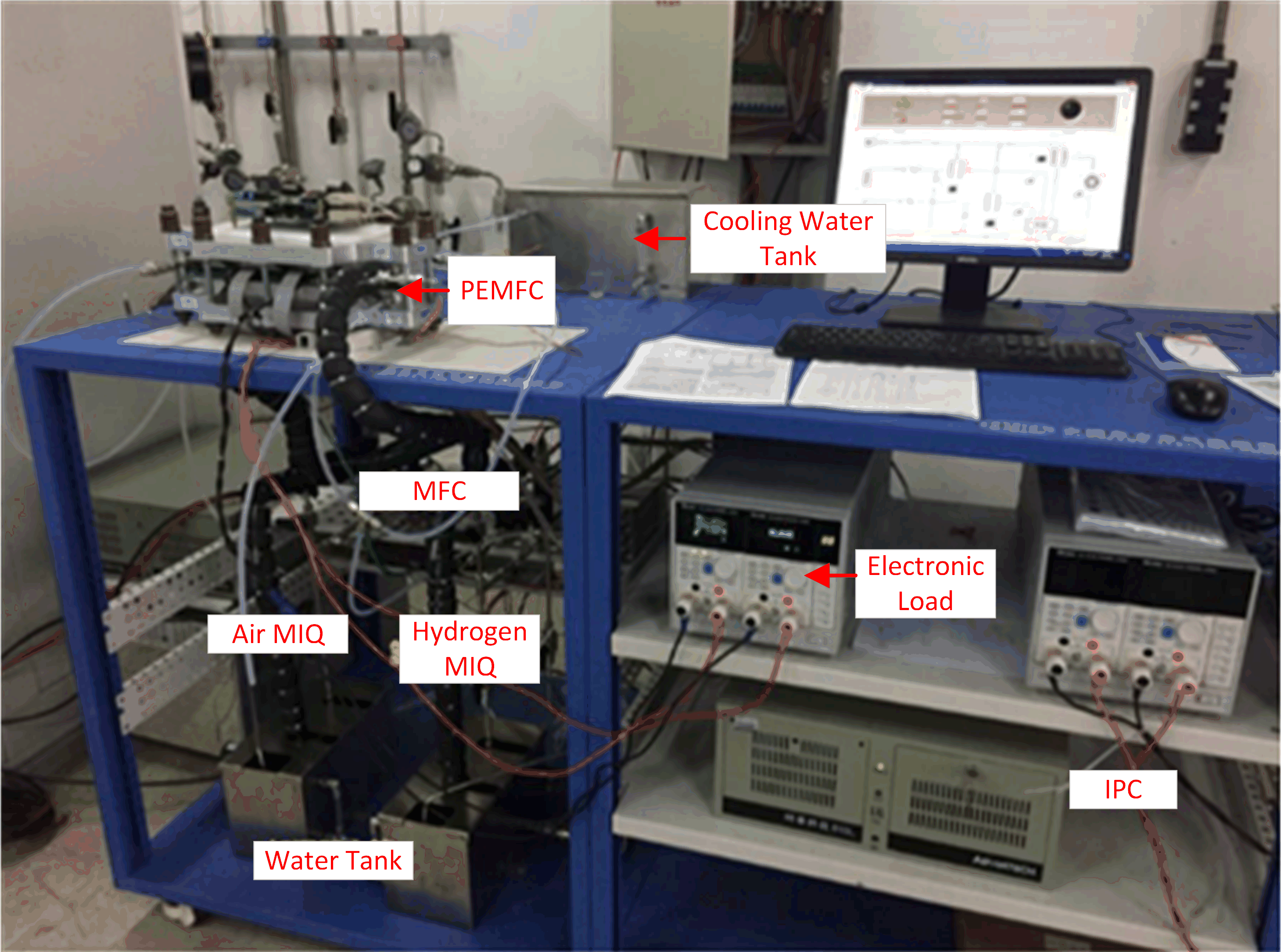}
	\caption{Experimental platform of the PEMFC.}
	\label{img_pemfc}
\end{figure}

\begin{table}[htbp]
  \centering
  \caption{Specifications of the PEMFC under normal condition}
    \begin{tabular}{cc}
    \toprule
    specification & parameter \\
    \midrule
    number of fuel cells & 6 \\
    rated load current/A & 80 \\
    sampling frequency/Hz & 10 \\
    air intake/splm & 2 \\
    hydrogen intake/splm & 1.8 \\
    rated temperature/℃ & 65 \\
    \bottomrule
    \end{tabular}%
  \label{normal_PEMFC}%
\end{table}%

\begin{figure}[htbp]
	\centering
		\includegraphics[scale=.35]{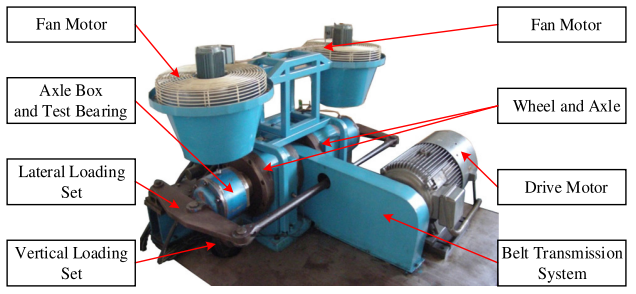}
	\caption{Experiment platform of WHELL.}
	\label{img_whell}
\end{figure}

\begin{table}[htbp]
  \centering
  \caption{Information about the fault states of the wheelset bearing dataset}
    \begin{tabular}{cc}
    \toprule
    label & state \\
    \midrule
    0 & normal \\
    1 & inner ring fault \\
    2 & rolling element fault \\
    3 & outer ring fult \\
    \bottomrule
    \end{tabular}%
  \label{info_whell}%
\end{table}%

\begin{table}[htbp]
  \centering
  \caption{Description of the four datasets used for validation}
    \begin{tabular}{cccccc}
    \toprule
    \textbf{dataset} & \textbf{dim} & \textbf{length} & \textbf{training} & \textbf{valid} & \textbf{testing} \\
    \toprule
    TE    & 52    & 100   & 58543 & 3082  & 10875 \\
    KDD-CUP99 & 39    & 100   & 72000 & 8000  & 20000 \\
    PEMFC & 6    & 30   & 27112 & 3000  & 6778 \\
    WHELL & 2    & 50   & 25650 & 2840  & 6420 \\
    \bottomrule
    \end{tabular}%
  \label{dasasets}%
\end{table}%

Details of the four time-series datasets used for performance comparison are described as follows.
\begin{enumerate}
\itemsep=0pt
\item The Tennessee-Eastman (TE) chemical process, witch simulated a real chemical production process. It has been widely used as a benchmark process to evaluate fault detection and fault diagnosis methods. This process includes one normal state and 20 fault states. Details of the TE process can be found in  \cite{downs1993plant}.
\item KDD-CUP99, witch recorded 7 million network traffic connection records within 7 weeks. The network traffic data were labeled as normal or attacked data. Additional data details can be found in \cite{lippmann20001999}.
\item Proton Exchange Membrane Fuel Cell (PEMFC), witch simulated a real fuel cell. As shown in Figure \ref{img_pemfc}, the PEMFC system is mainly composed of a fuel cell stack, an air supply subsystem, a hydrogen supply subsystem, a gas humidification subsystem, and a water management subsystem. The fuel cell of this experimental platform consists of 6 single cells, whose specifications under normal condition are shown in Table \ref{normal_PEMFC}. By simulating fault conditions artificially, the dataset formed 8 possible faults that may occur during operation, as listed in Table \ref{info_PEMFC}. The features of the PEMFC dataset are the voltages of the 6 cells in the cell stack and the sampling frequency is 10Hz.
\item Wheelset Bearing Dataset (WHELL), which simulated high-speed rail wheelset bearings. As shown in Figure \ref{img_whell}, the wheelset bearing test platform is mainly composed of a drive motor, a belt transmission system, a vertical loading set, a lateral loading set, two fan motors, and a control system. It simulates a real train using vertical and lateral loads. In this paper, this dataset is divided into four states, as shown in Table \ref{info_whell}. To simulate various complex working conditions of wheelset bearings during their operation as much as possible, under each health condition, vertical loads of 56, 146, 236, and 272 kN are set, and two lateral loads (0 and 20 kN) are set. The sampling frequency of the dataset is 5120Hz, and each sample contains two features, which are the data collected by the sensors \cite{wang2021feature}.
\end{enumerate}

The main information about the four datasets used in this paper is shown in Table \ref{dasasets}, including the feature dimensions, the length of the time series, and the numbers of samples in the training set, validation set and test set.

To prevent information leakage when processing datasets, this paper applies data non-crossover instead of sliding windows to collect time-series data. This means that each time node data point in the dataset would only belong to training samples or test samples. For fair comparison, we employ the same data preprocessing approach for all methods. 

\begin{table*}[htbp]
  \centering
  \caption{Network structures of the methods used for comparison.}
    \begin{tabular}{c|c}
    \toprule
    \textbf{Model} & \textbf{Structure} \\
    \midrule
    MLP   & View(x)→MLP(256,128,64)→softmax \\
    LSTM+MLP & LSTM(layer=2,hidden=128)→LSTM(layer=2,hidden=64)→MLP(256,128,64) →softmax \\
    BiLSTM+MLP & LSTM(layer=2,hidden=128)→LSTM(layer=2,hidden=64)→MLP(256,128,64) →softmax \\
    \multirow{2}[1]{*}{1DCNN+MLP} & 1DCNN(out\_channels=256,filter=5*5)→1DCNN(out\_channels=512,filter=7*7) \\
          & →1DCNN(out\_channels=256,filter=11*11) →MLP(256,128,64) →softmax \\
    \bottomrule
    \end{tabular}%
    \begin{tablenotes}    
        \item  All models use the same classification layer, MLP(256,128,64).       
      \end{tablenotes}
  \label{netStructure}%
\end{table*}%

\subsection{Models for comparison}
\label{mfc}

The proposed method is compared with the pure MLP, LSTM, BiLSTM and IDCNN models to verify its validity and efficiency. The network structures of the abovementioned method for comparison are shown in Table \ref{netStructure}. Since the core innovation of this paper is the proposed time relationship unit and decoupling position embedding unit that can be regarded as a new feature extractor, for fairness, the same classification layer is applied when comparing the performances of the methods. The structure of the MLP in the time relationship unit in the DPTRN is $ (512,128) $.

Furthermore, the DPTRN is compared with three advanced fault diagnosis methods, including FDGRU \cite{zhang2021fault}, MCNN-LSTM \cite{chen2021bearing}, and 1DCNN-VAF \cite{wang2021bearing}. 

\subsection{Implementation details}
\label{ID}

For fairness, all the compared models mentioned in this paper are trained and inferred on the same experimental platform (RTX 2080ti with 12G memory). The batch size of each network is set to $ 32 $, the epoch is $ 60 $, and the learning rate is $ 6e^-4 $. The L2 regularization coefficient is $ 0.0001 $.
\subsection{Case study}
\label{case1}

\begin{table*}[htbp]
  \centering
  \caption{The fault diagnosis results of various methods on the KDD-CUP99 dataset and TE dataset.}
  \resizebox{\textwidth}{!}{
    \begin{tabular}{ccccccc}
    \toprule
    \toprule
    \multirow{2}[4]{*}{\textbf{Method}} & \multicolumn{3}{c}{\textbf{KDD-CUP99 dataset}} & \multicolumn{3}{c}{\textbf{TE dataset}} \\
\cmidrule{2-7}      & \textbf{Recall(\%)} & \textbf{Accuracy (\%)} & \textbf{F1(\%)} & \textbf{Recall(\%)} & \textbf{Accuracy (\%)} & \textbf{F1(\%)} \\
    \midrule
    MLP & 0.9785(0.9762) & 0.9785(0.9762) & 0.9785(0.9762) & 0.9520(0.9434) & 0.9667(0.9650) & 0.9559(0.9468) \\
    LSTM+MLP & 0.9835(0.9830) & 0.9836(0.9829) & 0.9835(0.9829) & 0.9635(0.9226) & 0.9636(0.9180) & 0.9635(0.9128) \\
    BiLSTM+MLP & 0.9838(0.9830) & 0.9838(0.9830) & 0.9838(0.9830) & 0.9889(0.9338) & 0.9895(0.9208) & 0.9892(0.9218) \\
    1DCNN+MLP & \underline{0.9866}(0.9826) & \underline{0.9867}(0.9826) & \underline{0.9866}(0.9825) & 0.9898(0.9889) & 0.9904(0.9890) & 0.9901(0.9890) \\
    FDGRU \cite{zhang2021fault} & 0.9862(\underline{0.9844}) & 0.9862(\underline{0.9843}) & 0.9862(\underline{0.9844}) & \underline{0.9953}(\underline{0.9936}) & \underline{0.9956}(\underline{0.9938}) & \underline{0.9955}(\underline{0.9937}) \\
    1DCNN-VAF \cite{wang2021bearing} & 0.9657(0.9602) & 0.9641(0.9611) & 0.9648(0.9613)  & 0.9734(0.9705) & 0.9783(0.9751) & 0.9769(0.9749) \\
    MCNN-LSTM \cite{chen2021bearing} & 0.9829(0.9823) & 0.9830(0.9822) & 0.9829(0.9823) & 0.9856(0.9821) & 0.9880(0.9832) & 0.9869(0.9827) \\
    \textbf{DPTRN} & \textbf{0.9881(0.9870)} & \textbf{0.9881(0.9870)} & \textbf{0.9881(0.9870)} & \textbf{0.9960(0.9943)} & \textbf{0.9963(0.9948)} & \textbf{0.9961(0.9944)} \\
    \bottomrule
    \bottomrule
    \end{tabular}}
    \begin{tablenotes}    
        \item All models were trained on the same training set. Recall, accuracy, and F1 were all obtained on the same test set. The results outside the brackets are the best experimental results of 5 different random seeds. The results inside the brackets are the average results of 5 experiments.          
      \end{tablenotes}            
  \label{KDDresult}%
\end{table*}%

\begin{table*}[htbp]
  \centering
  \caption{The fault diagnosis results of various methods on the PEMFC dataset and WHELL dataset.}
  \resizebox{\textwidth}{!}{
    \begin{tabular}{ccccccc}
    \toprule
    \toprule
    \multirow{2}[4]{*}{\textbf{Method}} & \multicolumn{3}{c}{\textbf{PEMFC dataset}} & \multicolumn{3}{c}{\textbf{WHELL dataset}} \\
\cmidrule{2-7}      & \textbf{Recall(\%)} & \textbf{Accuracy (\%)} & \textbf{F1(\%)} & \textbf{Recall(\%)} & \textbf{Accuracy (\%)} & \textbf{F1(\%)} \\
    \midrule
    MLP & 0.9153(0.9107) & 0.9310(0.9246) & 0.9252(0.9118) & 0.9592(0.9581) & 0.9556(0.9536) & 0.9521(0.9507) \\
    LSTM+MLP & 0.9204(0.9180) & 0.9077(0.9019) & 0.9190(0.9125) & 0.9711(0.9692) & 0.9716(0.9703) & 0.9711(0.9698) \\
    BiLSTM+MLP & 0.9265(0.9230) & 0.9259(0.9231) & 0.9240(0.9216) & 0.9828(0.9810) & 0.9838(0.9825) & 0.9833(0.9827) \\
    1DCNN+MLP & 0.9322(0.9316) & 0.9275(0.9246) & 0.9272(0.9245) & 0.9882(0.9866) & 0.9884(0.9851) & 0.9847(0.9827) \\
    FDGRU \cite{zhang2021fault} & 0.9476(0.9453) & 0.9370(0.9333) & 0.9413(0.9291) & \textbf{0.9996(0.9986)} & \textbf{0.9996(0.9994)} & \textbf{0.9995(0.9992)} \\
    1DCNN-VAF \cite{wang2021bearing} & 0.9231(0.9204) & 0.9185(0.9116) & 0.9235(0.9225) & 0.9658(0.9639) & 0.9634(0.9617) & 0.9641(0.9630) \\
    MCNN-LSTM \cite{chen2021bearing} & \underline{0.9641}(\underline{0.9625}) & \underline{0.9669}(\underline{0.9617}) & \underline{0.9641}(\underline{0.9602}) & 0.9897(0.9877) & 0.9904(0.9887) & 0.9894(0.9882) \\
    \textbf{DPTRN} & \textbf{0.9836(0.9811)} & \textbf{0.9777(0.9764)} & \textbf{0.9828(0.9810)} & \underline{0.9912}(\underline{0.9916}) & \underline{0.9964}(\underline{0.9961}) & \underline{0.9943}(\underline{0.9939}) \\
    \bottomrule
    \bottomrule
    \end{tabular}}
  \label{PEMFC}%
\end{table*}%


In this section, we want to answer two questions:
\begin{enumerate}
\itemsep=0pt
\item Can the proposed DPTRN be guaranteed to be effective?
\item Can the proposed DPTRN have obvious advantages in computational efficiency?
\end{enumerate}

To answer the first question, the performances on the KDDCUP99 dataset (fault detection) and TE dataset, PEMFC dataset, WHELL dataset (fault diagnosis) are compared with those of the various baseline methods mentioned above in section \ref{The_effectiveness}. The detailed comparison results are shown in Table \ref{KDDresult} and Table \ref{PEMFC}. The best parts of the comparison results are bolded, and suboptimal results are underlined. The evaluation indicators used in the comparative experiment include the recall rate, accuracy rate and F1 value. 

To answer the second question, in section \ref{The_efficiency}, we compared the model size and the floating point operations (\textit{FLOPs}) of the DPTRN with those of various baseline methods to verify the efficiency of the proposed method. Model size measures the training difficulty and the complexity of the model. When the model size increases, the model will have more parameters, a better ability to fit the data, and more difficulty to converge. FLOPs measure the number of floating-point calculations required by the model to feed a sample forward, which is only related to the structure and operation of the model and is independent of hardware. FLOPs is a direct indicator of the computational efficiency of the model. The computational efficiency comparison results are shown in Table \ref{TEresult}.

\subsubsection{Effectiveness of different methods}
\label{The_effectiveness}

Table \ref{KDDresult} and Table \ref{PEMFC} show that the experimental performance of the simple MLP model is the worst. It is reasonable that it has a simple structure and does not have the ability to process a large number of features of the time-series data. In contrast, $ \rm{LSTM}+\rm{MLP} $ , $ \rm{BiLSTM}+\rm{MLP} $ and FDGRU achieve better fault detection or fault diagnosis results than simple MLP. This is because LSTM and GRU can use a recurring unit with shared weights to extract potential features and generate a hidden state vector to represent time evolution characteristics through serial calculation. The 1DCNN and 1DCNN-VAF also achieve satisfactory fault diagnosis performance because the 1DCNN uses different sized convolution kernels to interact features between different time nodes in time-series data. In addition, MCNN-LSTM, which combines the CNN and LSTM, also achieves good results.

Compared with various baseline methods, DPTRN achieves the best experimental results on the KDD-CUP99,TE and PEMFC datasets and suboptimal experimental results on the WHELL dataset. This is because the time relationship unit has powerful feature extraction capability on time-series data, and the decoupling position embedding unit can effectively synthesize the information of various time nodes, which indicates that our proposed DPTRN has the ability to effectively extract discriminating features on time-series data in a novel way.

Combining the above results on effectiveness, we can answer the first question mentioned earlier and state that the proposed DPTRN can perform fault detection or fault diagnosis effectively and has certain advantages in comparison with other baseline methods.


\subsubsection{Efficiency of different methods}
\label{The_efficiency}

\begin{table}[htbp]
  \centering
  \caption{Efficiency of various methods on the TE dataset.}
    \begin{tabular}{ccccccc}
    \toprule
    \toprule
    \multirow{2}[2]{*}{\textbf{Method}} & \multirow{2}[2]{*}{\textbf{Model size}} & \multirow{2}[2]{*}{\textbf{FLOPs}} \\
      &   &  \\
    \midrule
    MLP & 2995k & 1.37M \\
    LSTM+MLP & 368k & 31.19M \\
    BiLSTM+MLP & 368k & 85.27M \\
    1DCNN+MLP & 4679k & 91.88M \\
    FDGRU \cite{zhang2021fault} & 8473k & 637.92M \\
    1DCNN-VAF \cite{wang2021bearing} & 860k & 57.27M \\
    MCNN-LSTM \cite{chen2021bearing} & 332k & 120.28M \\
    \textbf{DPTRN} & \textbf{252k} & \textbf{17.24M} \\
    \bottomrule
    \bottomrule
    \end{tabular}%
  \label{TEresult}%
\end{table}%

To investigate the model efficiency, the model sizes and FLOPs of different methods are listed in Table \ref{TEresult}.

The simple MLP model has a large model size. This method uses the classification layer immediately after flattening the time-series data, which makes the input vector dimension extremely large and greatly increases the number of parameters. However, at the same time, due to the simple structure and parallel computing way of the simple MLP, it has few FLOPs, and the fault diagnosis task can be completed with increased efficiency. Since the recurrent units in LSTM and GRU share weights, the model size is greatly reduced, which reduces the difficulty of training. However, for LSTM, it is limited by its complex serial calculation design; therefore, the FLOPs of both $ \rm{LSTM}+\rm{MLP} $ and $ \rm{BiLSTM}+\rm{MLP} $ are significantly improved, which reduces their calculation efficiency. For FDGRU, it has the most parameters due to its complex structure, so it has the most FLOPs, which makes it the least efficient. The 1DCNN needs to use different convolution kernel sizes or deep structures to realize long-term feature interaction, which increases the model size and training difficulty of the CNN. In addition, although the CNN is computed in parallel, the complex convolution operation of the CNN will increase its number of FLOPs, so the 1DCNN is also unable to meet the requirement of real-time fault detection and fault diagnosis.

Our proposed DPTRN can extract features in parallel, which makes it more computationally efficient. As shown in Table \ref{TEresult}, the DPTRN has obvious advantages in model size, which makes training the DPTRN easier. In addition, because of its simple structure and the parallel computation method, the DPTRN has the fewest FLOPs, $ 55.27\% $, $ 20.21\% $, $ 18.76\% $, $ 2.7\% $, $ 30.11\% $ and $ 14.33\% $ of the number of FLOPs of the LSTM, BiLSTM, 1DCNN, FDGRU, 1DCNN-VAF and MCNN-LSTM models, respectively, which means that the proposed DPTRN has a very clear advantage in computational efficiency compared to the baseline methods. Especially when the number of FLOPs and model size of the proposed DPTRN are only $ 2.7\% $ and $ 2.97\% $ of those of the FDGRU, respectively, it surpasses the FDGRU on three datasets, and the results are very close on the WHELL dataset. The above results further prove that the proposed method has a sufficiently sensitive response for fault diagnosis and can meet the increasing need for high efficiency in industrial processes. Considering the results of each baseline method, the parallel computing method has fewer FLOPs and higher computational efficiency, which verifies the effectiveness of parallel computing emphasized in this paper.

Combining the above results on efficiency, we can answer the second question mentioned earlier, that the proposed DPTRN has a significant advantage in computational efficiency compared to other baseline methods.

Based on the results obtained on the above four datasets, the DPTRN method proposed in this paper has the advantage of high computational efficiency due to its parallel computing characteristics. In addition, it has fewer parameters and is easier to train the model because its time relationship unit shares weights and the model can be summed as a 6-layer MLP structure. Furthermore, the DPTRN model achieves the best fault diagnosis and fault detection results on three datasets and one suboptimal results on the WHELL dataset, which further proves the effectiveness, rationality and robustness of the time relationship unit and decoupling position embedding unit proposed in this paper. Therefore, our proposed DPTRN can realize efficient and reliable fault diagnosis and has the potential to meet the real-time fault detection or fault diagnosis requirement in industrial processes.

\begin{table*}[htbp]
  \centering
  \caption{Ablation experiment results of the DPTRN on the KDDCUP99 and TE datasets}
  \resizebox{\textwidth}{!}{
    \begin{tabular}{ccccccc}
    \toprule
    \toprule
    \multirow{2}[4]{*}{\textbf{Method}} & \multicolumn{3}{c}{\textbf{KDD-CUP99 dataset}} & \multicolumn{3}{c}{\textbf{TE dataset}} \\
\cmidrule{2-7}      & \textbf{Recall(\%)} & \textbf{Accuracy (\%)} & \textbf{F1(\%)} & \textbf{Recall(\%)} & \textbf{Accuracy (\%)} & \textbf{F1(\%)} \\
    \midrule
    $ \rm{DPTRN}_a $ & 0.9813(0.9785) & 0.9813(0.9807) & 0.9813(0.9791) & 0.9825(0.9795) & 0.9858(0.9839) & 0.9830(0.9805) \\
    $ \rm{DPTRN}_b $ & 0.9752(0.9713) & 0.9752(0.9758) & 0.9752(0.9723) & 0.9741(0.9314) & 0.9800(0.9590) & 0.9754(0.9340) \\
    \textbf{DPTRN} & \textbf{0.9881(0.9870)} & \textbf{0.9881(0.9870)} & \textbf{0.9881(0.9870)} & \textbf{0.9960(0.9943)} & \textbf{0.9963(0.9948)} & \textbf{0.9961(0.9944)} \\
    \bottomrule
    \bottomrule
    \end{tabular}}%
    \begin{tablenotes}    
        \item $ \rm{DPTRN}_a $ does not use the position embedding unit but only uses the time relationship unit. $ \rm{DPTRN}_b $ is the model that uses the absolute position embeddings instead of using the decoupling position embedding unit while using the time relationship unit.          
      \end{tablenotes}
  \label{Ablation}%
\end{table*}%

\begin{table*}[htbp]
  \centering
  \caption{Ablation experiment results of the DPTRN on the PEMFC and WHELL datasets}
  \resizebox{\textwidth}{!}{
    \begin{tabular}{ccccccc}
    \toprule
    \toprule
    \multirow{2}[4]{*}{\textbf{Method}} & \multicolumn{3}{c}{\textbf{PEMFC dataset}} & \multicolumn{3}{c}{\textbf{WHELL dataset}} \\
\cmidrule{2-7}      & \textbf{Recall(\%)} & \textbf{Accuracy (\%)} & \textbf{F1(\%)} & \textbf{Recall(\%)} & \textbf{Accuracy (\%)} & \textbf{F1(\%)} \\
    \midrule
    $ \rm{DPTRN}_a $ & 0.9773(0.9755) & 0.9711(0.9707) & 0.9753(0.9741) & 0.9774(0.9741) & 0.9836(0.9817) & 0.9794(0.9787) \\
    $ \rm{DPTRN}_b $ & 0.9692(0.9681) & 0.9678(0.9668) & 0.9681(0.9673) & 0.9687(0.9659) & 0.9752(0.9728) & 0.9723(0.9710) \\
    \textbf{DPTRN} & \textbf{0.9836(0.9811)} & \textbf{0.9777(0.9764)} & \textbf{0.9828(0.9810)} & \textbf{0.9912(0.9916)} & \textbf{0.9964(0.9961)} & \textbf{0.9943(0.9939)} \\
    \bottomrule
    \bottomrule
    \end{tabular}}%
  \label{Ablation_2}%
\end{table*}%
\subsection{Ablation study}
\label{case2}

\begin{figure*}[htbp]
	\centering
		\includegraphics[scale=0.7]{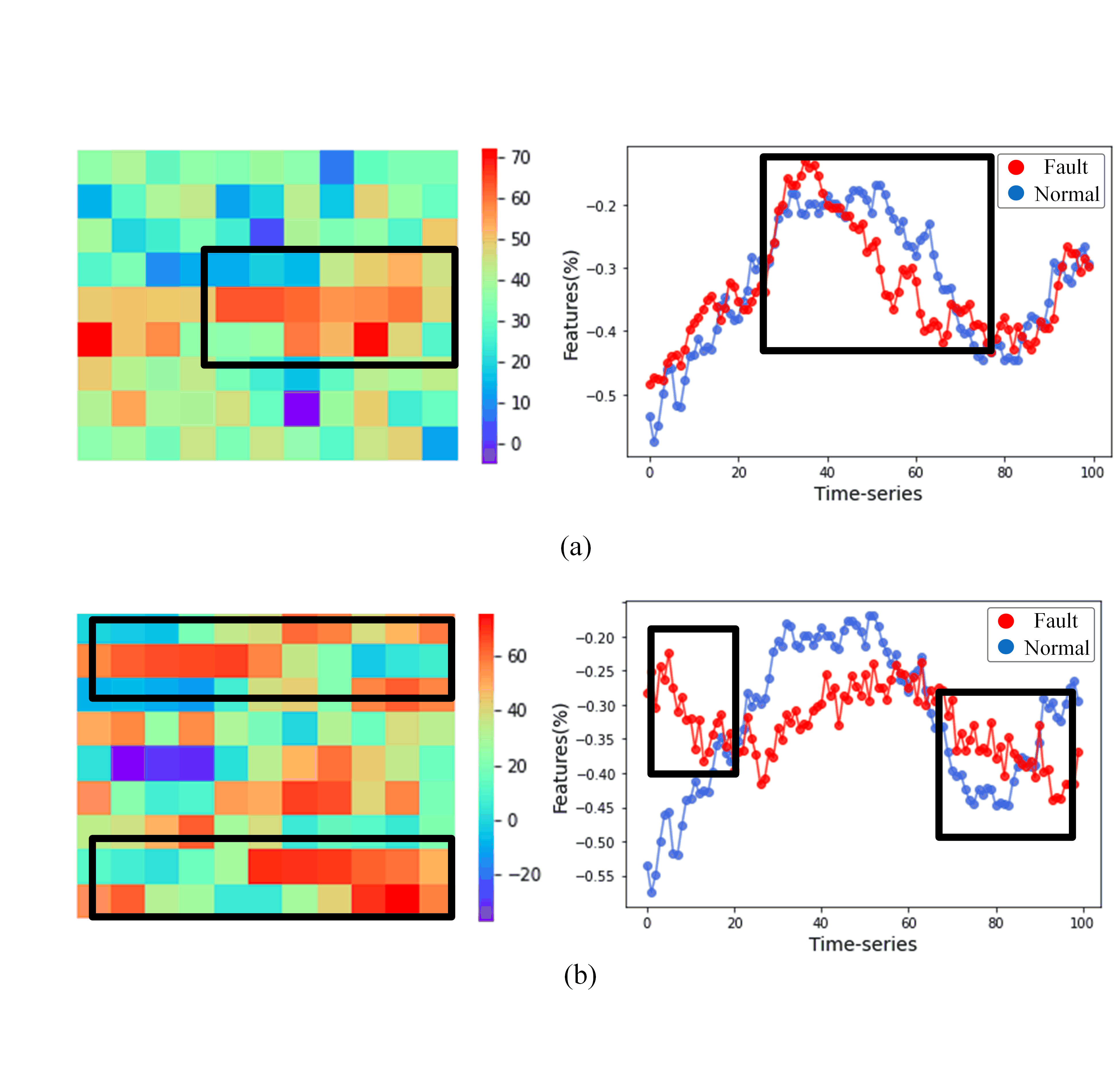}	\caption{Heatmap output of the time relationship unit and the corresponding feature representation for the TE dataset. (a) and (b) show the results for the two fault samples, respectively. There are 2 subgraphs in the figure. The left half of each subgraph is the heatmap output, which represents the preliminary relationship weight output of the time relationship unit. Each block represents a time node from 0 to T-1, and the figure is arranged from left to right and from top to bottom. The right half is the feature of each time node. The blue curve indicates the normal sample, and the red curve indicates the fault sample. The data masked with a black box indicate the part showing a significant difference between normal and fault samples.}
	\label{img4}
\end{figure*}

To further verify the validity and rationality of the proposed time relationship unit and the decoupling position embedding unit separately. we conducted ablation experiments. For $ \rm{DPTRN}_a $, it does not use the position embedding unit but only uses the time relationship unit (that is, it directly uses the preliminary relationship weight in Formula \ref{RWPRE} as the final relationship weight). For $ \rm{DPTRN}_b $, the model in Figure \ref{img2}, only uses the absolute position embeddings instead of using the decoupling position embedding unit while using the time relationship unit. In order to distinguish, the model proposed in this paper is called DPTRN in the comparison. The experimental results are shown in Table \ref{Ablation} and Table \ref{Ablation_2}.

All of the metrics of $ \rm{DPTRN}_a $ on the four datasets are more than 97\% and surpass or are close to those of the simple MLP, LSTM,  MCNN-LSTM and 1DCNN-VAF models. This demonstrates the effectiveness of the time relationship unit, which is better able to extract the characteristics of time-series data.

The performance metrics of $ \rm{DPTRN}_b $ are not very satisfying, and even worse than those of the simple MLP on the TE dataset. This is consistent with the hypothesis that the absolute position embeddings cannot be directly applied in fault diagnosis because it introduces noise to the raw data.

Our proposed DPTRN model is significantly better than $ \rm{DPTRN}_a $ and $ \rm{DPTRN}_b $, which further verifies the advantages of the framework effectiveness of the decoupling position embedding unit proposed in this paper.

The results of the above ablation experiments show that the time relationship unit can effectively extract the evolution features of the time-series data, and the decoupling position embedding unit proposed in this paper can further improve the performance of the model.

\subsection{Interpretability of the DPTRN}
\label{case3}

The proposed framework combined with the time relationship unit and the decoupling position embedding unit can significantly improve the performance, so investigating how the two units affect the results is of interest. Therefore, we randomly selected two fault samples from the TE dataset and compared them with normal samples to check the output of the two units, and analyze the internal working principles of these two units.

\subsubsection{Interpretability of the time relationship unit}


\begin{figure}[htpb]
	\centering
		\includegraphics[scale=.5]{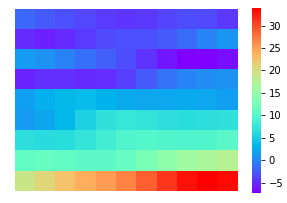}
	\caption{Weight output of the decoupling position embedding unit in the TE dataset. According to the time node from 0 to T-1, the figure is arranged from left to right and from top to bottom.}
	\label{img5}
\end{figure}

\begin{figure*}[htpb]
	\centering
		\includegraphics[scale=0.7]{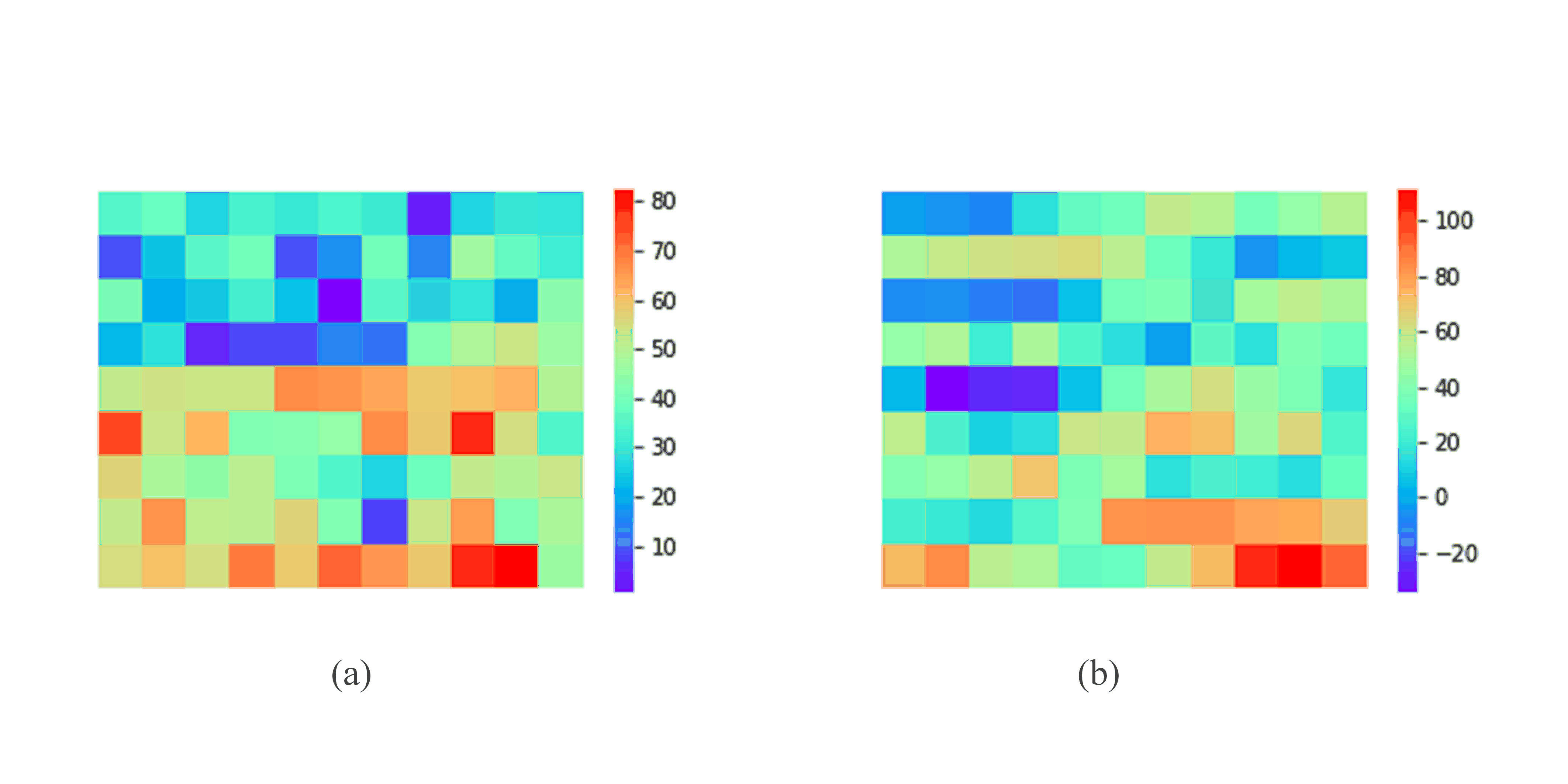}
	\caption{Relationship weight heatmaps of the two fault samples output from the combination of the two units. According to the time node from 0 to T-1, the figure is arranged from left to right and from top to bottom. The darker red cubes indicate that the current time nodes have a higher preliminary relationship weight, and the darker blue cubes show the opposite.}
	\label{img6}
\end{figure*}

For the time relationship unit, the outputs of the two fault samples mentioned above are shown in Figure \ref{img4}. In the heatmap of each subgraph, the darker red cubes indicate that the current time nodes have a higher preliminary relationship weight, and the darker blue cubes indicate the opposite. For the left subgraph, the outputs with larger weights are dark red. Integrating the left subgraph with the right subgraph shows that the red parts map the features that indicate the significant difference between the fault sample and normal sample. The corresponding parts of the two subgraphs are marked with black boxex and these marked time nodes contain long-term evolutionary information that is useful for correct fault diagnosis.

In addition, the time relationship unit outputs completely different preliminary relation weights for the two types of fault samples. For each sample, historical time nodes that contain important information for fault diagnosis may be distributed anywhere in the length of the time series, which requires the time relationship unit to output the correct weight at each time node. By using the different weights to extract the evolution information in the time-series data, the subsequent classification layer can more easily focus on these historical nodes that are beneficial to fault diagnosis, which greatly improves the performance of fault diagnosis.

\subsubsection{Interpretability of the decoupling position embedding unit}

Since the input of the decoupling position embedding unit is the absolute position embedding determined by the timing length, the outputs of each sample are the same and shown in Figure \ref{img5}. The weight outputs of the decoupling position embedding unit increase as the historical time node approaches the current time node. Therefore, for the decoupling position embedding unit, the closer the distance to the current time node, the more valuable the time nodes are. 

It is worth noting that the outputs of the decoupling position embedding unit are learnable, which means that it has learned results that are consistent with human prior knowledge, that is, the reference value of information from a long time ago will decrease. This result shows that the decoupling position embedding unit works as we expect, that is, it introduces the context information into the model without introducing noise to the raw data.

\subsubsection{Interpretability of the relationship weight}

For the two random time series fault samples mentioned above, the final outputs obtained by combining the time relationship unit and the decoupling position embedding unit are shown in Figure \ref{img6}. As shown in the figure, the final relationship weights become smoother, and have the ability to indicate both the importance of each historical time node and the relative importance of the times nodes among the time series for fault diagnosis.

Neural network structures, such as RNN, LSTM and GRU, use a hidden state vector to represent the features of the evolution of the time-series data, which cannot explain the specific physical meaning. CNN apply convolution kernels to complete the feature interaction of time-series data, and it is also difficult to explain the specific physical meaning of the convolution kernel. For our proposed DPTRN, the time relationship unit explains the importance of each historical time node for fault diagnosis, and the decoupling position embedding unit explains the relationship of the importance of historical information over time.

Therefore, the proposed DPTRN is more interpretable than the traditional deep neural network.

\section{Conclusion}
\label{conclu}
This work has been submitted to the IEEE for possible publication. Copyright may be transferred without notice, after which this version may no longer be accessible.

In this article, a deep parallel time-series relation network model for fault diagnosis of time-series data is proposed. Considering that traditional deep learning networks are limited to parallel computing when analyzing time-series data, the DPTRN model proposed in this paper uses serial time relationship units to extract evolutionary features and improve computational efficiency. To introduce the contextual information of time-series data to the model, a decoupling position embedding unit is introduced to the DPTRN.

Fault diagnosis and fault detection experiments were carried out on the TE, KDD-CUP99, PEMFC and WHELL datasets. The experimental results show that the proposed DPTRN model has better effectiveness on three datasets. It is worth discussing why our proposed DPTRN does not achieve the best results on the WHELL dataset. We think there are two reasons. First, as seen from Table \ref{TEresult}, the FDGRU has the most parameters, so it has a better data fitting ability than the DPTRN. Second, our proposed DPTRN relies on the feature interaction between different time nodes for feature extraction, and the feature dimension of WHELL data is small, with 2 dimensions, which reduces the feature extraction performance of the DPTRN. Nevertheless, our proposed DPTRN still achieves more than 99\% on the three metrics, 2.77\% higher than 1DCNN-VAF and 0.4\% higher than MCNN-LSTM in average recall. In addition, we compared robustness and computational efficiency; our DPTRN has the best computational efficiency. In addition, the output of the time relation unit and the decoupling position embedding unit is analyzed in detail, and the results show the reliability and stronger interpretability of the DPTRN.

When dealing with time-series data, this paper refocuses attention on MLP and proves the feasibility and effectiveness of parallel computing. This provides a new idea for solving the problem of time-series data fault diagnosis. 

In this paper, only the original features of historical time nodes are used, while the label information of each historical node is ignored. Further research on how to fully use historical data will be an interesting direction. Furthermore, how to effectively interact features with smaller dimensions deserves further research.

\bibliographystyle{unsrt}  
\bibliography{references}

\begin{thebibliography}{10}

\bibitem{xu2018novel}
Yuan Xu, Sheng-Qi Shen, Yan-Lin He, and Qun-Xiong Zhu.
\newblock A novel hybrid method integrating ica-pca with relevant vector
  machine for multivariate process monitoring.
\newblock {\em IEEE Transactions on Control Systems Technology},
  27(4):1780--1787, 2018.

\bibitem{zhang2019novel}
Jian Zhang, Jianxiao Zou, Jiyang Zhang, Qian Tao, Xingtai Gui, Hongbing Xu, and
  Shicai Fan.
\newblock A novel deep dpca-svm method for fault detection in industrial
  processes.
\newblock In {\em 2019 IEEE 58th Conference on Decision and Control (CDC)},
  pages 2916--2921. IEEE, 2019.

\bibitem{yang2021fault}
Chun Yang, Lujing Tao, Jian Zhang, Xingtai Gui, Jiyang Zhang, Jianxiao Zou, and
  Shicai Fan.
\newblock A fault detection method based on the deep extended pca--svm in
  industrial processes.
\newblock In {\em 2021 American Control Conference (ACC)}, pages 3620--3625.
  IEEE, 2021.

\bibitem{wei2020research}
Yuqin Wei and Zhengxin Weng.
\newblock Research on te process fault diagnosis method based on dbn and
  dropout.
\newblock {\em The Canadian Journal of Chemical Engineering}, 98(6):1293--1306,
  2020.

\bibitem{wang2020novel}
Yalin Wang, Zhuofu Pan, Xiaofeng Yuan, Chunhua Yang, and Weihua Gui.
\newblock A novel deep learning based fault diagnosis approach for chemical
  process with extended deep belief network.
\newblock {\em ISA transactions}, 96:457--467, 2020.

\bibitem{hoang2019survey}
Duy-Tang Hoang and Hee-Jun Kang.
\newblock A survey on deep learning based bearing fault diagnosis.
\newblock {\em Neurocomputing}, 335:327--335, 2019.

\bibitem{cabrera2020bayesian}
Diego Cabrera, Adriana Guam{\'a}n, Shaohui Zhang, Mariela Cerrada, Rene-Vinicio
  Sanchez, Juan Cevallos, Jianyu Long, and Chuan Li.
\newblock Bayesian approach and time series dimensionality reduction to
  lstm-based model-building for fault diagnosis of a reciprocating compressor.
\newblock {\em Neurocomputing}, 380:51--66, 2020.

\bibitem{zhang2019bidirectional}
Shuyuan Zhang, Kexin Bi, and Tong Qiu.
\newblock Bidirectional recurrent neural network-based chemical process fault
  diagnosis.
\newblock {\em Industrial \& Engineering Chemistry Research}, 59(2):824--834,
  2019.

\bibitem{lavrova2019using}
Daria Lavrova, Dmitry Zegzhda, and Anastasiia Yarmak.
\newblock Using gru neural network for cyber-attack detection in automated
  process control systems.
\newblock In {\em 2019 IEEE International Black Sea Conference on
  Communications and Networking (BlackSeaCom)}, pages 1--3. IEEE, 2019.

\bibitem{zollanvari2020transformer}
Amin Zollanvari, Kassymzhomart Kunanbayev, Saeid~Akhavan Bitaghsir, and Mehdi
  Bagheri.
\newblock Transformer fault prognosis using deep recurrent neural network over
  vibration signals.
\newblock {\em IEEE Transactions on Instrumentation and Measurement}, 70:1--11,
  2020.

\bibitem{canizo2019multi}
Mikel Canizo, Isaac Triguero, Angel Conde, and Enrique Onieva.
\newblock Multi-head cnn--rnn for multi-time series anomaly detection: An
  industrial case study.
\newblock {\em Neurocomputing}, 363:246--260, 2019.

\bibitem{gu2021data}
Xin Gu, Zhongjun Hou, and Jun Cai.
\newblock Data-based flooding fault diagnosis of proton exchange membrane fuel
  cell systems using lstm networks.
\newblock {\em Energy and AI}, 4:100056, 2021.

\bibitem{wang2021intelligent}
Hongwei Wang, Wenlei Sun, Li~He, and Jianxing Zhou.
\newblock Intelligent fault diagnosis method for gear transmission systems
  based on improved multi-scale reverse dispersion entropy and swarm
  decomposition.
\newblock {\em IEEE Transactions on Instrumentation and Measurement}, 71:1--13,
  2021.

\bibitem{chen2021comparative}
Zhiwen Chen, Ketian Liang, Steven~X Ding, Chao Yang, Tao Peng, and Xiaofeng
  Yuan.
\newblock A comparative study of deep neural network-aided canonical
  correlation analysis-based process monitoring and fault detection methods.
\newblock {\em IEEE Transactions on Neural Networks and Learning Systems},
  2021.

\bibitem{nie2021two}
Xiaoyin Nie and Gang Xie.
\newblock A two-stage semi-supervised learning framework for fault diagnosis of
  rotating machinery.
\newblock {\em IEEE Transactions on Instrumentation and Measurement}, 70:1--12,
  2021.

\bibitem{wang2021feature}
Huan Wang, Zhiliang Liu, Dandan Peng, Mei Yang, and Yong Qin.
\newblock Feature-level attention-guided multitask cnn for fault diagnosis and
  working conditions identification of rolling bearing.
\newblock {\em IEEE Transactions on Neural Networks and Learning Systems},
  2021.

\bibitem{xu2019online}
Gaowei Xu, Min Liu, Zhuofu Jiang, Weiming Shen, and Chenxi Huang.
\newblock Online fault diagnosis method based on transfer convolutional neural
  networks.
\newblock {\em IEEE Transactions on Instrumentation and Measurement},
  69(2):509--520, 2019.

\bibitem{huang2021fault}
Deqing Huang, Shupan Li, Na~Qin, and Yuanjie Zhang.
\newblock Fault diagnosis of high-speed train bogie based on the
  improved-ceemdan and 1-d cnn algorithms.
\newblock {\em IEEE Transactions on Instrumentation and Measurement}, 70:1--11,
  2021.

\bibitem{liu2021pay}
Hanxiao Liu, Zihang Dai, David~R So, and Quoc~V Le.
\newblock Pay attention to mlps.
\newblock {\em arXiv preprint arXiv:2105.08050}, 2021.

\bibitem{tolstikhin2021mlp}
Ilya Tolstikhin, Neil Houlsby, Alexander Kolesnikov, Lucas Beyer, Xiaohua Zhai,
  Thomas Unterthiner, Jessica Yung, Daniel Keysers, Jakob Uszkoreit, Mario
  Lucic, et~al.
\newblock Mlp-mixer: An all-mlp architecture for vision.
\newblock {\em arXiv preprint arXiv:2105.01601}, 2021.

\bibitem{lee2021fnet}
James Lee-Thorp, Joshua Ainslie, Ilya Eckstein, and Santiago Ontanon.
\newblock Fnet: Mixing tokens with fourier transforms.
\newblock {\em arXiv preprint arXiv:2105.03824}, 2021.

\bibitem{vaswani2017attention}
Ashish Vaswani, Noam Shazeer, Niki Parmar, Jakob Uszkoreit, Llion Jones,
  Aidan~N Gomez, {\L}ukasz Kaiser, and Illia Polosukhin.
\newblock Attention is all you need.
\newblock In {\em Advances in neural information processing systems}, pages
  5998--6008, 2017.

\bibitem{ioffe2015batch}
Sergey Ioffe and Christian Szegedy.
\newblock Batch normalization: Accelerating deep network training by reducing
  internal covariate shift.
\newblock In {\em International conference on machine learning}, pages
  448--456. PMLR, 2015.

\bibitem{downs1993plant}
James~J Downs and Ernest~F Vogel.
\newblock A plant-wide industrial process control problem.
\newblock {\em Computers \& chemical engineering}, 17(3):245--255, 1993.

\bibitem{lippmann20001999}
Richard Lippmann, Joshua~W Haines, David~J Fried, Jonathan Korba, and Kumar
  Das.
\newblock The 1999 darpa off-line intrusion detection evaluation.
\newblock {\em Computer networks}, 34(4):579--595, 2000.

\bibitem{zhang2021fault}
Yahui Zhang, Taotao Zhou, Xufeng Huang, Longchao Cao, and Qi~Zhou.
\newblock Fault diagnosis of rotating machinery based on recurrent neural
  networks.
\newblock {\em Measurement}, 171:108774, 2021.

\bibitem{chen2021bearing}
Xiaohan Chen, Beike Zhang, and Dong Gao.
\newblock Bearing fault diagnosis base on multi-scale cnn and lstm model.
\newblock {\em Journal of Intelligent Manufacturing}, 32:971--987, 2021.

\bibitem{wang2021bearing}
Xin Wang, Dongxing Mao, and Xiaodong Li.
\newblock Bearing fault diagnosis based on vibro-acoustic data fusion and
  1d-cnn network.
\newblock {\em Measurement}, 173:108518, 2021.

\end{thebibliography}

\end{document}